\documentclass[10pt,twocolumn, letterpaper]{article}  %

\usepackage{cvpr}              %
\usepackage[accsupp]{axessibility}  %
\usepackage{graphicx}
\usepackage{amsmath}
\usepackage{amssymb}
\usepackage{algorithm}
\usepackage{algpseudocode}
\usepackage{booktabs}
\usepackage{blindtext}
\usepackage{overpic}
\usepackage{xspace}
\usepackage{colortbl}
\usepackage{xcolor}
\usepackage{pifont}
\usepackage{lipsum}
\usepackage{tikz}
\usepackage{microtype}
\usetikzlibrary{shadows.blur}

\definecolor{turquoise}{cmyk}{0.65,0,0.1,0.3}
\definecolor{purple}{rgb}{0.65,0,0.65}
\definecolor{dark_green}{rgb}{0, 0.3, 0}
\definecolor{orange}{rgb}{1, 0.65, 0}
\definecolor{red}{rgb}{0.8, 0.2, 0.2}
\definecolor{darkred}{rgb}{0.6, 0.1, 0.05}
\definecolor{blueish}{rgb}{0.0, 0.3, .6}
\definecolor{light_gray}{rgb}{0.7, 0.7, .7}
\definecolor{pink}{rgb}{0.9, 0, 0.6}
\definecolor{greyblue}{rgb}{0.25, 0.25, 1}
\definecolor{teal}{rgb}{0.0, 0.4, 0.4}
\definecolor{chocolate}{rgb}{1.0, 0.4, 0.0}
\definecolor{dark_blue}{rgb}{0.26, 0.43, 0.97}

\definecolor{amr_color}{rgb}{0.85,0.0,0.55}

\usepackage{soul}
\setuldepth{foobar}

\renewcommand{\paragraph}[1]{\vspace{.5em}\noindent\textbf{#1}.}

\DeclareMathOperator*{\argmin}{arg\,min}

\newcommand{\losst}[1]{\mathcal{L}_\text{#1}}
\newcommand{\real}{\mathbb{R}}
\newcommand{\ray}{\mathbf{r}}

\newcommand{\radiance}{\mathbf{c}}

\newcommand{\density}{\sigma}
\newcommand{\transmittance}{T}

\newcommand{\identity}{\mathbf{f}}
\newcommand{\prob}{\mathbf{p}}
\newcommand{\labl}{\mathbf{y}}

\newcommand{\primal}{\mathbf{p}}

\newcommand{\cell}{\mathbf{c}}

\newcommand{\neighbors}{\mathcal{N}}
\newcommand{\facearea}{\mathbf{A}}

\newcommand{\pixelidentity}{\mathbf{I}}
\newcommand{\numsegclass}{\mathcal{K}}

\definecolor{color1}{rgb}{0.9, 0.65, 0.65}
\definecolor{color2}{rgb}{0.95, 0.8, 0.8}
\definecolor{color3}{rgb}{1.0, 0.9, 0.9}
\newcommand{\best}[1]{\hspace{-\fboxsep}\colorbox{color1}{\textbf{#1}}\hspace{-\fboxsep}}
\newcommand{\second}[1]{\hspace{-\fboxsep}\colorbox{color2}{\textit{#1}}\hspace{-\fboxsep}}

\definecolor{cvprblue}{rgb}{0.21,0.49,0.74}
\usepackage[pagebackref,breaklinks,colorlinks,allcolors=cvprblue]{hyperref}

\def\paperID{5507} %
\def\confName{CVPR}
\def\confYear{2026}

\title{Semantic Foam: Unifying Spatial and Semantic Scene Decomposition}

\author{
{Amr Sharafeldin}$^{1}$\quad
{Shrisudhan Govindarajan\thanks{Equal advising} }$^{1}$\quad
{Thomas Walker}$^{5}$\quad
{Aryan Mikaeili}$^{1}$\quad 
 \\
{Daniel Rebain}$^{3}$\quad
{Kwang Moo Yi}$^{4}$\quad 
{Andrea Tagliasacchi\footnotemark[1] }$^{1,2,3}$\quad 
\\[0.5em]
$^1$Simon Fraser University \quad
$^2$University of Toronto \quad
$^3$Wayve Technologies \\
$^4$University of British Columbia \quad
$^5$University of Edinburgh\quad
}

\begin{document}
\maketitle

\begin{abstract}
Modern scene reconstruction methods, such as 3D Gaussian Splatting, deliver photo-realistic novel view synthesis at real-time speeds, yet their adoption in interactive graphics applications has been limited. 
A major bottleneck is the difficulty of interacting with these representations compared to traditional, human-authored 3D assets. 
While previous research has attempted to impose semantic decomposition on these models, significant challenges remain regarding segmentation quality and consistency. 
To address this, we introduce Semantic Foam, extending the recently proposed Radiant Foam representations to semantic decomposition tasks.
Our approach integrates the natural spatial volumetric decomposition of Radiant Foam's Voronoi mesh with an explicit semantic feature field parameterized at the cell level. 
This explicit structure enables direct spatial regularization, which prevents artifacts caused by occlusion or inconsistent supervision across views -- common pitfalls for other point-based representations. 
Experimental results show that our method achieves comparable or superior object-level segmentation performance compared to state-of-the-art methods like Gaussian Grouping and SAGA.
Project page: \url{http://semanticfoam.github.io/} 
\end{abstract}
    
\begin{figure}[t]
\includegraphics[width=\linewidth]{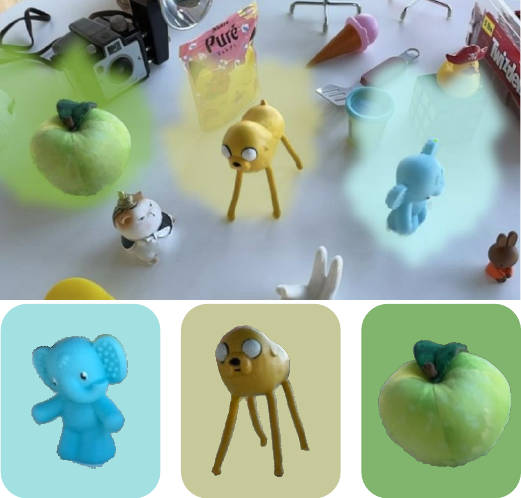}
\caption{{\bf Teaser -- }
We propose \textit{Semantic Foam}, a semantically decomposed 3D representation for scenes.
Based on the Radiant Foam~\cite{govindarajan2025radfoam} model, our method extends its spatial Voronoi decomposition to also separate space into semantically distinct regions.
This decomposition is regularized to extend into the empty space immediately surrounding objects (top), which enables clean extraction (bottom) and insertion edits that neither leave floating artifacts behind, nor miss parts of isolated objects, as is often the case for Gaussian-based semantic decompositions.
}
\label{fig:teaser}
\end{figure}

\section{Introduction}
\label{sec:intro}

Modern techniques for 3D scene reconstruction are capable of building detailed and photorealistic representations of the world from images.
However, many of the best performing methods produce reconstructions consisting of many millions of unstructured primitives, which are difficult to process in downstream applications due to a lack of semantic understanding.
An ideal reconstruction method should~(as illustrated in ~\Cref{fig:teaser}) produce models which are both \textit{spatially} and \textit{semantically} decomposed, such that edits and processing can be done in a way that is aware of both the physical and logical structure of objects in the scene.

While the problem of spatial decomposition can be addressed through modifications of the representation and training strategy~\cite{govindarajan2025radfoam}, semantic decomposition in 3D poses a unique challenge: training data, in the form of ground truth sematic volumetric labeling of space, is scarce.
Therefore, any model which aims to predict this must derive this information by other means.
This is typically achieved by leveraging 2D semantic labels or features~(i.e., images), which can now be readily generated cheaply and with reasonable quality by large computer vision foundation models.

Various methods have been proposed to achieve semantic decomposition in volumetric models based on neural fields~\cite{zhi2021semantic, mirzaei2022laterf, kobayashi2022distilledfeaturefields, isrfgoel2023} using the approach of supervising from 2D semantic features or labels.
However, the practical usage of such models for tasks that require editing is limited by the continuous nature of the representation -- 
decomposing a neural field into distinct physical regions is challenging, and requires post-processing heuristics.

Point-based representations like Gaussian Splatting~\cite{kerbl3Dgaussians} have also been used with this approach by associating each primitive with semantic information~\cite{cen2023segment, zhang2025labelgs, jain2024gaussiancutinteractivesegmentationgraph, cen2023saga, zhao2025isegmaninteractivesegmentandmanipulate3d}.
However, these methods must contend with the fact that Gaussian-based representations are not ``truly'' volumetric, resulting in unlabeled primitives in object interiors, as well as other artifacts caused by a lack of semantic connectivity.
As an example, Gaussian Grouping~\cite{gaussian_grouping} computes the convex hull of removed objects to avoid leaving behind mislabeled Gaussians, which can result in sections of unrelated geometry also being removed, as shown in \Cref{fig:gg_failure}.
This also restricts the model to extracting only convex objects, thereby limiting its application.
Overlapping primitives can also lead to contradicting semantic signals for the same 3D point, resulting in semantic ambiguity.

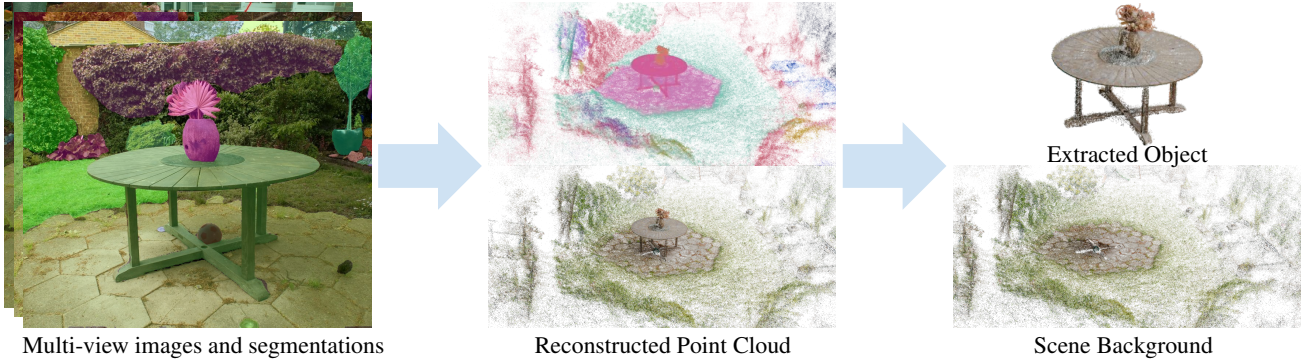
\begin{figure*}
    \centering
    \begin{overpic}[width=\linewidth]{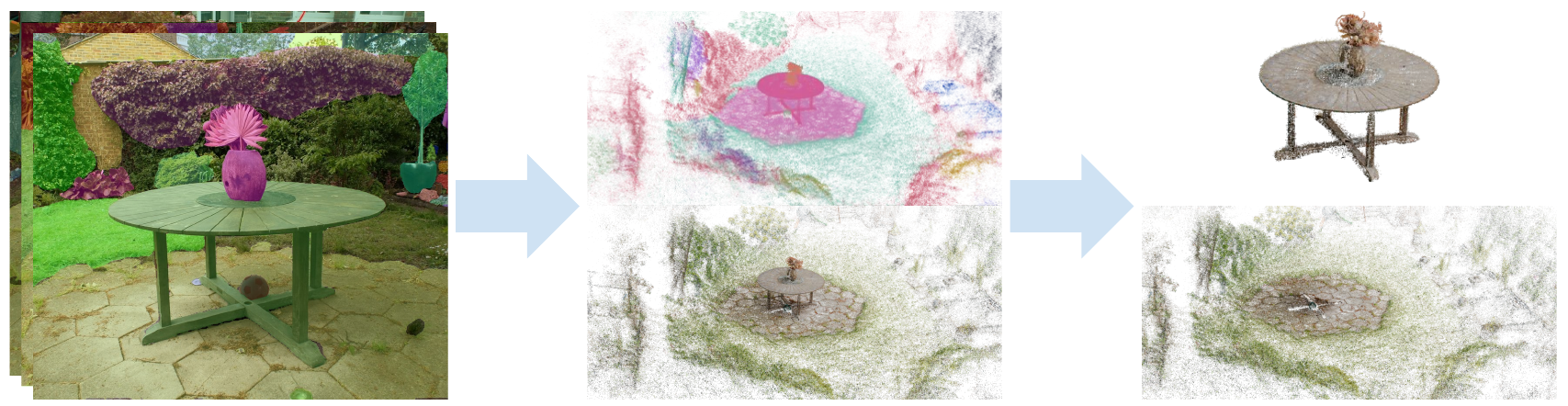}
    \put(2, -1.5){\small{Multi-view images and segmentations}}
    \put(41, -1.5){\small{Reconstructed Point Cloud}}
    \put(79, -1.5){\small{Scene Background}}
    \put(80, 13){\small{Extracted Object}}
    \end{overpic}
        \vspace{0.25em}
    \captionof{figure}{
        {\bf Overview -- }
        Our method builds on Radiant Foam, adding an extra supervision channel in the form of segmentation masks predicted by image segmentation models.
        Using these masks alongside the original images (left), Semantic Foam constructs a volumetric mesh-based radiance field along with per-point semantic identity features (center).
        Using these semantic features we can perform editing operations like object extraction and deletion (right).}
        \label{fig:subteaser}
\end{figure*}

To overcome these limitations, we adopt the recently introduced Radiant Foam~\cite{govindarajan2025radfoam} representation, augment it with per-primitive semantic information, and supervise it with 2D DEVA masks~\cite{cheng2023deva}.
Radiant Foam combines the point-based advantages of Gaussian Splatting methods with the volumetric nature of NeRF models.
It also provides explicit connectivity between the cells of its mesh structure, which can be leveraged to learn spatially smooth and consistent semantic decompositions.
Our \textit{Semantic Foam} model does this by applying a total variation objective that keeps the semantic identity features of nearby cells consistent, even when they lack direct supervision from image space features due to occlusion.
The resulting scene reconstructions are both spatially decomposed (defining an inside/outside relation for all points in space), and semantically decomposed (defining a semantic label for all points in space).
The intersection of these decomposition enables clean editing, insertion, and extraction of objects without leaving behind excess mislabeled primitives and floating artifacts.

We demonstrate several potential applications of our method, including object removal, extraction and, insertion.
On object-level segmentation, our method matches or outperforms Gaussian-based segmentation approaches like Gaussian Grouping~\cite{gaussian_grouping} and SAGA~\cite{cen2023saga} by 2–6\% in mIoU across three complex, real-world datasets.

\vspace{.5em}
\noindent
To summarize our main contributions:
\vspace{0em} 
\begin{itemize}
    \item We extend Radiant Foam's volumetric mesh representation to support label-aware 3D segmentation. 
    \item We introduce a novel volumetric total variation loss based on the Voronoi adjacency graph for enforcing spatial smoothness and feature consistency.
    \item We demonstrate the editing potential of our approach, including object extraction, removal, and insertion in complex 3D scenes.
\end{itemize}

\section{Related Work}

3D segmentation is a long-standing problem, with early works focusing on geometric and topological cues to segment 3D meshes into semantically meaningful parts~\cite{shamir2006seg, katz2003hierarchical}.
With the introduction of large-scale annotated 3D datasets, the field shifted toward learning-based methods, operating on point clouds~\cite{qi2017pointnet, qi2017pointnetplusplus, thomas2019KPConv}, voxel grids~\cite{graham2018sparse, choy20194d, simonelli2025easy3d}, and meshes~\cite{hanocka2019meshcnn, kalogerakis2010seg}.
While these approaches achieve strong results on curated datasets, they offer limited flexibility and interactivity,  rely on dense 3D supervision, and remain limited in generalizing to open-world scenes where geometric and semantic boundaries are less well-defined.

\paragraph{Foundation models for 3D segmentation}
Recently, large-scale vision foundation models such as SAM~\cite{kirillov2023segany} and SAMv2~\cite{ravi2024sam} have enabled fine-grained, open-world segmentation without the need for task-specific finetuning.
Their strong generalization allows obtaining dense, class-agnostic masks for arbitrary objects, which can be leveraged for 3D scene understanding by lifting 2D segmentations or distilling their features into 3D through multi-view consistency or differentiable rendering~\cite{lang2024iseg, chen2023bridging, abdelreheem2023zeroshot}.
Beyond purely visual models, multimodal foundation models such as CLIP \cite{radford2021clip}, DINO~\cite{caron2021emerging,oquab2023dinov2}, and recent text-to-image diffusion models~\cite{rombach2021highresolution} have been utilized to couple visual and semantic embeddings for open-vocabulary scene understanding~\cite{lerf2023, perla2025asia, ha2022semabs}.
These models provide rich, transferable priors that significantly reduce the need for curated 3D annotations, marking a shift from supervised segmentation toward foundation-guided 3D representation learning.

\paragraph{NeRF segmentation}
The advent of Neural Radiance Fields~\cite{mildenhall2020nerf} introduced a powerful paradigm for 3D scene reconstruction from multi-view images, representing geometry and appearance as continuous volumetric fields.
Their differentiable volumetric rendering formulation enables lifting features from 2D images into a consistent 3D space, allowing semantic cues from pretrained 2D models to be integrated across views.
Building on this property, several works extended NeRFs toward semantic and panoptic scene understanding. 
For example, NVOS~\cite{ren-cvpr2022-nvos} learns a voxelized feature field supervised with coarse 2D scribbles and uses 3D graph-cut for segmentation.
SemanticNeRF~\cite{zhi2021semantic} and LaTeRF~\cite{mirzaei2022laterf} extend NeRF by introducing a dedicated probability branch alongside the appearance and geometry fields, leveraging the continuity of the volumetric representation to propagate sparse 2D labels into a consistent 3D semantic space.
Distilled Feature Fields ~\cite{kobayashi2022distilledfeaturefields} and ISRF~\cite{isrfgoel2023} further generalizes this idea by adding a feature branch that distills representations from pretrained 2D vision models such as DINO~\cite{caron2021emerging} into the 3D field, enabling scene segmentation and editing.
LERF~\cite{lerf2023} tackles text-based object selection by distilling CLIP features of image crops into the 3D representation. While this enables open-vocabulary reasoning, the global nature of CLIP features can make the precise localization of object boundaries difficult.
OpenNeRF~\cite{engelmann2024opennerf} addresses this limitation by leveraging OpenSeg~\cite{ghiasi2021open} features, which provide more spatially localized semantics. 
Similarly, GARField~\cite{garfield2024} enhances localization by supervising the 3D field with multi-scale SAM masks, effectively decoupling spatial scales to achieve sharper segmentation boundaries.

\paragraph{3DGS segmentation}
Despite their success in semantic scene understanding, NeRF-based methods remain limited by their slow training and rendering speed. 
3D Gaussian Splatting (3DGS)~\cite{kerbl3Dgaussians} has emerged as an efficient alternative, enabling real-time rendering by modeling the scene as a collection of anisotropic Gaussian kernels.
Extending the capabilities of 3DGS to model semantics, SA3D~\cite{cen2023segment} assigns a label to each Gaussian in the scene, classifying them as either background or foreground, and supervises this process using masks obtained from SAM. 
Similarly, LabelGS~\cite{zhang2025labelgs} extends this formulation by lifting multi-class pixel labels to individual Gaussians for fine-grained scene segmentation.
GaussianCut~\cite{jain2024gaussiancutinteractivesegmentationgraph} performs background–foreground segmentation using a graph-cut formulation guided by coarse user-provided scribbles. 
SAGA~\cite{cen2023saga} distills SAM features directly into each Gaussian, enabling interactive query-based segmentation. 
The iSegMan method~\cite{zhao2025isegmaninteractivesegmentandmanipulate3d} introduces a training-free, voting-based algorithm that assigns pixel segments to Gaussians for localized, text-driven editing. 
FeatureSplatting~\cite{qiu-2024-featuresplatting} and LangSplat~\cite{qin2023langsplat} further extend these ideas by distilling DINO and CLIP features into Gaussians, facilitating language-guided object extraction and selection.
Most similar to our approach, Gaussian Grouping~\cite{gaussian_grouping} augments each Gaussian with a learnable feature vector, which is subsequently passed through a linear layer for multi-class classification supervised using SAM-generated masks.

While 3DGS offers fast rendering and supports localized segmentation, the continuous nature of Gaussian kernels and their approximate representation of scene geometry hinder the formation of sharp and well-defined boundaries. 
Moreover, the unstructured spatial layout of Gaussians makes neighborhood-based regularization computationally expensive~\cite{gaussian_grouping}. 
To overcome these limitations, we adopt Radiant Foams~\cite{govindarajan2025radfoam} as the underlying 3D representation. 
Its piecewise-constant volumetric formulation naturally produces crisp segmentation boundaries, while the inherent adjacency graph of the volumetric cells enables efficient and spatially consistent regularization.

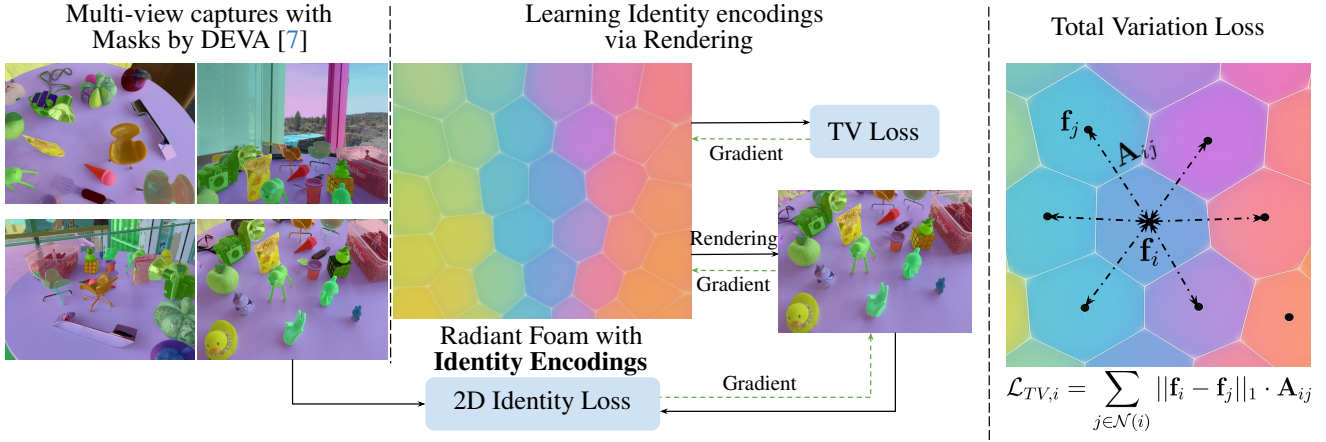
\begin{figure*}
    \centering
    \begin{overpic}[width=\linewidth]{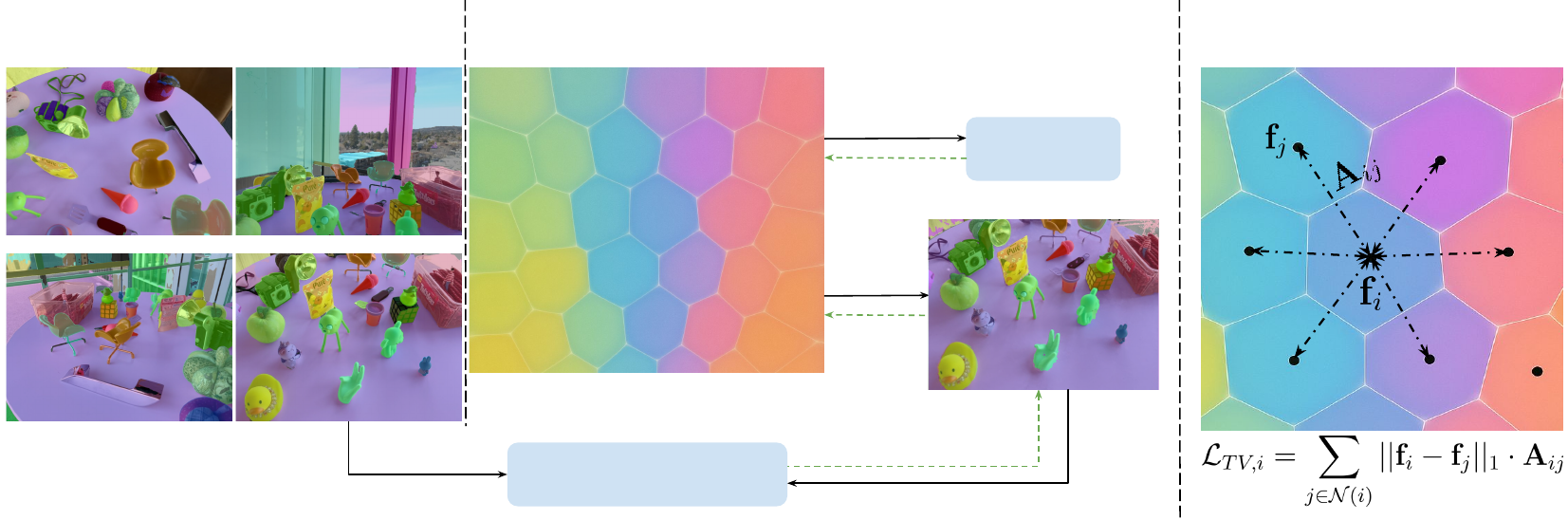}
    \put(5, 32){{Multi-view captures with}}
    \put(7, 30){{Masks by DEVA~\cite{cheng2023deva}}}
    \put(40, 32){{Learning Identity encodings}}
    \put(46, 30){{via Rendering}}
    \put(80, 31){{Total Variation Loss}}
    \put(34.5, 2.5){{2D Identity Loss}}
    \put(63, 23){{TV Loss}}
    \put(33.5, 7.5){{Radiant Foam with}}
    \put(33, 5.5){{\textbf{Identity Encodings}}}
    \put(55, 4){{\footnotesize{Gradient}}}
    \put(53, 11.5){{\footnotesize{Gradient}}}
    \put(52.5, 15){{\footnotesize{Rendering}}}
    \put(54, 21.5){{\footnotesize{Gradient}}}
    \end{overpic}
    \captionof{figure}{
        {\bf Semantic Foam Training -- }
        The training pipeline of Semantic Foam consists of two primary stages: 
        (left) we begin by preparing the inputs using DEVA in everything mode to automatically generate masks for all training views; 
        (middle) given these masked multi-view images, we jointly optimize all properties of the 3D Voronoi cells—including their identity encodings -- via differentiable rendering, supervised by an Identity Loss and a Total Variation (TV) Loss; 
        (right) we illustrate the formulation of the TV Loss, which minimizes the L1 difference between the identity encodings of each Voronoi site and its neighboring cells.
        }
        \vspace{1em}
        \label{fig:method}
\end{figure*}

\section{Method}
We start by reviewing the Radiant Foam representation, which establishes the foundation for our work. 
We then present our primary contribution: an extension that integrates object identity capture directly into the Radiant Foam representation. 
Finally, we demonstrate how these learned semantic segments enable intuitive, post-processing-free scene editing capabilities.

\subsection{Radiant Foam (preliminaries)}
Radiant Foam models a 3D scene using a non-overlapping, differentiable volumetric mesh derived from a Voronoi diagram. 
This geometric construction enables constant-time ray traversal and efficient rendering. 
The Voronoi diagram partitions the domain into convex polyhedral cells, $\mathcal{C}_i$. 
Each cell~$\mathcal{C}_i$ is defined as the set of all points $\mathbf{x}$ in $\mathbb{R}^3$ that are closer to its associated primal vertex $\mathbf{p}_i$ (a site of the diagram) than to any other site $\mathbf{p}_j$:
\begin{align}
    \cell_i = \{x \in \real^3 : \argmin_j ||x - \primal_j|| = i\}.\label{eq:voronoi}
\end{align}
The volumetric radiance field within these cells is parameterized by constant density and view-dependent appearance~(represented as spherical harmonics).
This parameterization is subsequently utilized for volume rendering to compute the final radiance along a ray:
\begin{align}
\radiance_\ray &= \sum_{n=1}^{N} \transmittance_n \cdot (1 - \textrm{exp}(-\density_n \delta_n)) \cdot \radiance_n, \label{eq:volrendeq}\\
\transmittance_n &= \prod_{j=1}^n \textrm{exp}(-\density_j \delta_j),
\end{align}
where $\delta_n$ is the width of segment $n$.
Although Radiant Foam provides an effective spatial partitioning for scene representation, its formulation is not directly applicable to scene editing. 
This limitation stems from the model's exclusive focus on appearance and geometry, neglecting the semantic categorization required for targeted scene manipulation. 
We address this by proposing a modification to Radiant Foam: learning 3D consistent object identity, which is supervised using dense 2D mask proposals from Decoupled Video Segmentation Approach (DEVA)~\cite{cheng2023deva}.

\subsection{Semantic Foam}
To equip Radiant Foam with semantic capabilities, we introduce Semantic Foam. 
This extension preserves all existing geometric and photometric attributes of the Voronoi cells~(position, density, and spherical harmonics coefficients) to maintain high-fidelity scene representation. 
We augment each cell with a new \textit{Identity Encoding} feature, which enables the model to represent 3D semantic instances or object identities.
The Identity Encoding Feature is parameterized as a \textit{piecewise constant}, \textit{view-independent} vector assigned to each cell. 
This design choice is grounded in the physical realization that 3D semantic information should not depend on the camera's viewing direction.

To train these Identity Encoding Features, we compute a differentiable semantic identity rendering ($\pixelidentity_\ray$) along each ray similar to the standard appearance rendering process. 
This process integrates the Identity Encoding Feature ($\identity$) along the ray using the volume rendering formulation:
\begin{align}
    \pixelidentity_\ray &= \sum_{n=1}^{N} \transmittance_n \cdot (1 - \textrm{exp}(-\density_n \delta_n)) \cdot \identity_n. \label{idrendeq}
\end{align}
The resulting vector $\pixelidentity_\ray$ represents the projection of the 3D-consistent semantic representation into image space. 
During the training phase, this rendered vector is mapped to a set of $\numsegclass$ semantic identity classes through a single linear layer:
\begin{align}
    \prob_k = softmax_k(\text{LL}(\pixelidentity_\ray))
\end{align}
where $\text{LL}(\cdot)$ denotes a linear layer mapping semantic identity vector to class probabilities $\prob_k$ that the ray belongs to class $k$. 
We optimize these classification logits against ground-truth segmentation masks in a joint training objective with the native photometric loss used in Radiant Foam. 
This unified optimization ensures that Semantic Foam maintains strict geometric integrity while achieving high semantic coherence.

\subsection{Loss Function}
\label{loss_function}

To train the identity features to capture 3D-consistent semantic identities, we utilize two losses, which we now describe.

\paragraph{2D Identity loss} 
Since the Ground Truth (GT) semantic mask identities are available in image space, we supervise the identity encoding features through the rendered identity features of the ray $\pixelidentity_\ray$ with the ground truth semantic segmentations. 
During training, the rendered identity vector is mapped into $\numsegclass$ semantic identity logits using a single linear layer.
The logits are then transformed into class probabilities $\prob_k$ for each class $k$ using the softmax function.
These resulting semantic class probabilities are subsequently optimized with the ground truth segmentation masks using the cross-entropy loss:
\begin{align}
  \losst{identity} &= \sum_{k=1}^{\numsegclass} -\labl_k \textrm{log}(\prob_k),
\end{align}
where $\labl_k$ is the one-hot ground-truth label in image space.

\paragraph{Total-variation loss}
While Radiant Foam’s volumetric mesh provides a structural basis for 3D segmentation, supervising identity solely through the 2D identity loss presents a significant limitation. 
Specifically, training only with the identity loss frequently results in a failure to enforce smooth segmentation boundaries and generates noisy identity features, thereby limiting their practical utility.
Regions that contribute minimally to the rendered image -- such as object interiors -- receive insufficient supervision, as the gradient magnitude for a cell is proportional to its volume rendering weight. 
This lack of constraint allows interior cells to learn spurious identity features, resulting in noisy decompositions that hinder downstream editing. 
To address this, we leverage the explicit mesh adjacency of the Voronoi diagram to implement a total-variation (TV) loss. 
We define a weighted $\losst{1}$ loss between the identity feature $\identity_i$ of a cell and the identity features $\identity_j$ of its adjacent neighbors. 
The loss is weighted by the surface area of the shared face between cells, effectively regularizing the identity field to be spatially smooth and suppressing noise in unobserved regions.
\begin{align}
    \losst{TV,i} &= \sum_{j \in \neighbors(i)} || \identity_i - \identity_j ||_1 \cdot \facearea_{ij},
\end{align}
where $\neighbors(i)$ denotes the set of cells that share a face with $\cell_i$, and $\facearea_{ij}$ represents the surface area of the mutual face shared by cells $\cell_i$ and $\cell_j$. 
The aggregate total-variation loss is calculated as the mean of $\losst{TV,i}$ across all cells $i$.

\subsection{Optimization}
The optimization protocol strictly adheres to the strategy established in the original Radiant Foam. 
Training commences with initialization using a sparse point cloud derived from COLMAP~\cite{schoenberger2016sfm}.
Throughout the optimization procedure, we utilize the same densification and pruning heuristics from Radiant Foam: new Voronoi sites are introduced in under-represented areas, and unutilized cells are removed. 
During densification, the identity encoding feature for any newly created Voronoi cell is cloned from its parent, consistent with the treatment of color and density parameters.

\paragraph{Training objective} 
The training objective integrates the native rendering loss $\losst{rgba}$ and quantile loss $\losst{quantile}$, which are necessary for learning scene appearance and geometry in the original Radiant Foam framework. 
Additionally, we augment these with the semantic identity loss $\losst{identity}$ and the total-variation loss $\losst{TV}$ to enable the joint optimization of the semantic encoding features.
The overall training objective for our model is defined as:
\begin{align}
    \losst{} &= \underbrace{\losst{rgba} + \lambda_{1} \losst{quantile}}_{\text{Radiant Foam}} + \underbrace{\lambda_{2} \losst{identity} + \lambda_{3} \losst{TV}}_{\text{Semantic Foam}}.
\end{align}
To generate the required input ground truth masks, we follow the procedure proposed by LabelGS~\cite{zhang2025labelgs}, which employs a strategy of generating densified novel views for input to DEVA~\cite{cheng2023deva} in order to generate masks with better cross-view consistency.

\subsection{Implementation Details}
We extend the Radiant Foam codebase by incorporating volume rendering of identity encodings, enabling pixel-level identity supervision. 
This required modifying the custom CUDA kernels to support both forward and backward rendering passes for the identity field. Across all datasets, we employ a 16-dimensional identity encoding for semantic understanding.
$\quad$
Following Radiant Foam's training procedure, we train each scene for 20k iterations and adopt the same densification and pruning strategy for adaptive allocation of Voronoi sites. 
For Mip-NeRF360~\cite{mipnerf360} and LERF-Mask~\cite{gaussian_grouping}, we set the identity-loss weight to $\lambda_3 {=} 1000$, while for LLFF~\cite{llff} we use $\lambda_3 {=} 300$. 
The total variation loss weight is exponentially decayed by a factor of $0.99$ every $1000$ iterations beginning at iteration $2000$, which reduces oversmoothing and preserves spatial detail in the identity encodings. 

\subsection{Semantic Foam for Scene editing}
\label{subsec:editing}
A fully trained Semantic Foam provides a semantically aware 3D representation. 
Prior to inference, the semantic class for each cell is determined by passing its identity encoding feature through the learned linear layer. 
Our resulting representation facilitates a range of downstream scene editing tasks, including object extraction, removal, and insertion, without requiring expensive post-processing.

\paragraph{Object Extraction}
Given a target semantic class, we first select all high-density cells with semantic identity same as the target class. 
In the Radiant Foam representation, however, this initial selection is insufficient for full extraction because object surfaces are inherently defined by the interface between high-density cells and empty (near-zero density) space. 
To obtain the proper object boundary -- the visible ``shell" -- we must additionally extract the 1-ring neighbors of the high-density cells, as defined by the explicit Voronoi diagram adjacency. 
By leveraging this adjacency to include neighboring empty cells that share the same semantic identity, we ensure the object surface is fully extracted.

\paragraph{Object Removal}
Object removal follows a similar workflow. 
Given a target semantic class, we compute the volumetric mask using the extraction method described above. 
These identified cells are then pruned from the representation. 
Finally, the Voronoi diagram is re-triangulated to maintain a continuous, non-overlapping partition of the remaining space, effectively completing the removal operation.

\paragraph{Object Insertion}
To insert a new object at a particular location, we first delete any existing Voronoi cells occupying that spatial region. 
The primal vertices of the new object are then introduced into the representation, and the mesh is re-triangulated to seamlessly integrates the object into the global scene representation.

\section{Experiments}

\begin{table}[t]
\centering
\resizebox{\linewidth}{!}{
\setlength{\tabcolsep}{2pt}
\begin{tabular}{l|cc|cc|cc}
    \toprule
    & \multicolumn{2}{c|}{MipNeRF 360~\cite{mipnerf360}} & \multicolumn{2}{c|}{LERF-Masked~\cite{gaussian_grouping}} & \multicolumn{2}{c}{LLFF~\cite{llff}} \\
    & mIoU$\uparrow$ & mAcc$\uparrow$ & mIoU$\uparrow$ & mAcc$\uparrow$ & mIoU$\uparrow$ & mAcc$\uparrow$ \\
    \midrule 
    LabelGS~\cite{zhang2025labelgs}           & 0.70          & \second{0.94} & 0.55& \second{0.95} & 0.64& 0.75\\
    Gaussian Gouping~\cite{gaussian_grouping} & 0.75& 0.87& \second{0.77}& 0.86& \best{0.86}& \best{0.91}\\
    SAGA~\cite{cen2023saga}                   & \second{0.81} & \best{0.97}   & 0.71& \best{0.96}& 0.73& 0.81\\
    \textbf{SemanticFoam}                     & \best{0.82}   & 0.94& \best{0.78}& 0.88& \second{0.85}& \second{0.90}\\
    \bottomrule
\end{tabular}
}
\caption{
    {\bf Segmentation --}
    We evaluate our method's accuracy in segmentation on held-out test views for three standard datasets. 
    Our method achieves competitive or superior mIoU performance in comparise to Gaussian-based segmentation methods.
}
\label{tab:quant_res}
\end{table}

To evaluate segmentation accuracy in open-world scenarios, we conduct experiments across a comprehensive suite of 16 scenes from three established public datasets. 
Specifically, we assess our method on five scenes from Mip-NeRF 360~\cite{mipnerf360} (excluding the bicycle and stump scenes due to significant label inconsistencies in the DEVA~\cite{cheng2023deva} output), three scenes from LERF-Mask~\cite{gaussian_grouping}, and eight scenes from LLFF~\cite{llff}.

The data was pre-processed to filter out occluded background segmentation masks, ensuring a focus on primary scene geometry. 
To ensure a rigorous comparison, we adhere to the salient‑object selection protocol established by LabelGS~\cite{zhang2025labelgs} and generate object masks for all test‑time camera views accordingly. 
We benchmark our approach against state-of-the-art baselines, including LabelGS~\cite{zhang2025labelgs}, Gaussian Grouping~\cite{gaussian_grouping}, and SAGA~\cite{cen2023saga}, for both qualitative and quantitative assessment.

\paragraph{Metrics} 
We assess each method using two widely recognized segmentation quality metrics: mean Intersection over Union (mIoU) and mean Accuracy (mAcc). 
In the supplementary material, we provide extensive visual comparisons of segmentation masks across novel video trajectories, demonstrating the robustness of our method under significant viewpoint variations.

\begin{figure*}
    \centering
    \begin{overpic}[width=0.9\linewidth]{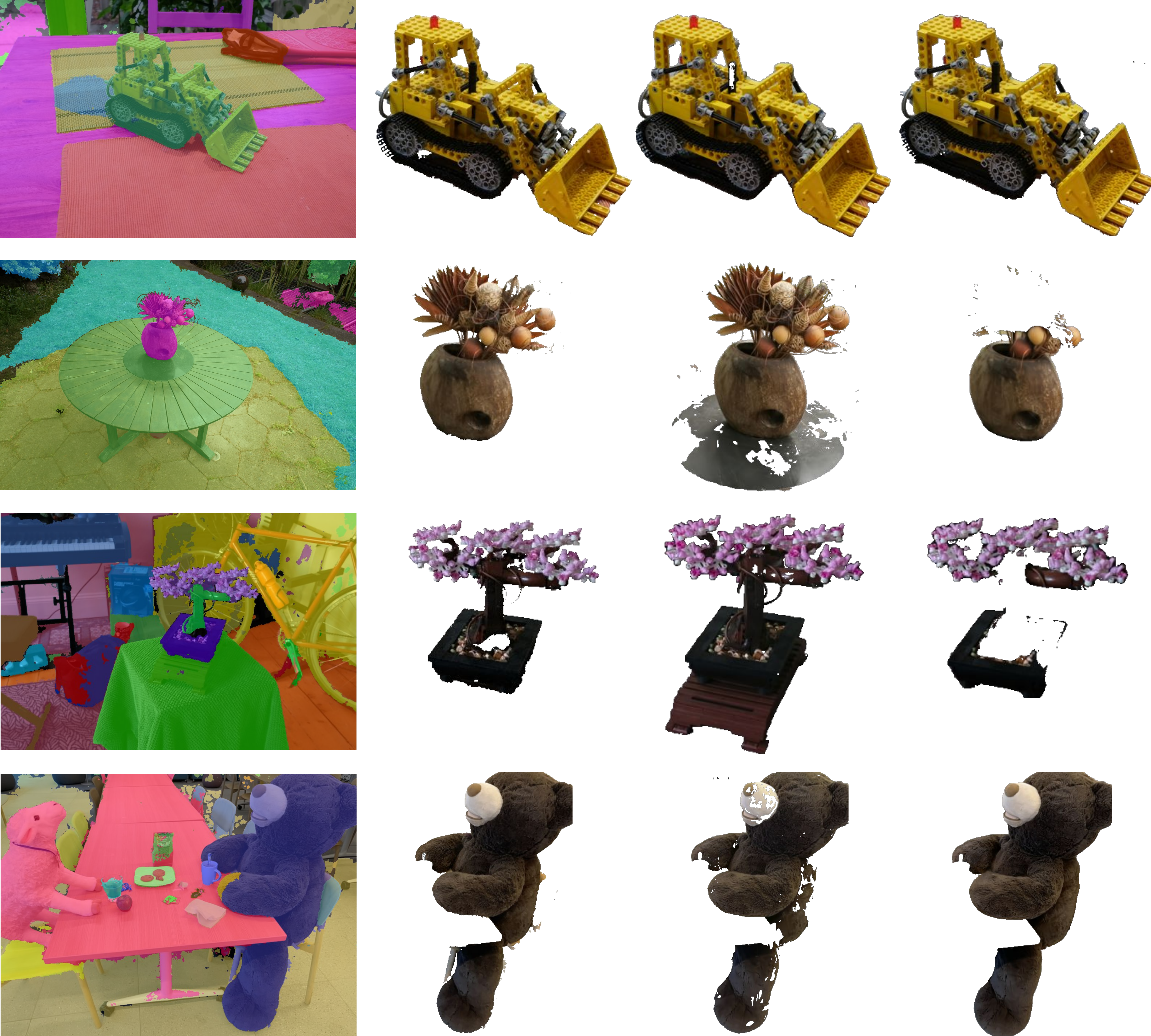}
    \put(-3, 75){\rotatebox{90}{Kitchen~\cite{mipnerf360}}}
    \put(-3, 53){\rotatebox{90}{Garden~\cite{mipnerf360}}}
    \put(-3, 32){\rotatebox{90}{Bonsai~\cite{mipnerf360}}}
    \put(-3, 7){\rotatebox{90}{Teatime~\cite{lerf2023}}}
    \put(6, 92){Our segmentation}
    \put(36, 92){Ours}
    \put(58, 92){SAGA~\cite{cen2023saga}}
    \put(77, 92){Gaussian Grouping~\cite{gaussian_grouping}}
    \end{overpic}
    \caption{
    {\bf Qualitative results -- }
    We present qualitative comparisons of our object‑extraction results against Gaussian Grouping~\cite{gaussian_grouping} and SAGA~\cite{cen2023saga}. 
    As illustrated by the extracted pot and leaves, Gaussian‑based approaches frequently over‑ or under‑segment object regions, whereas our method produces precise, well‑bounded object masks that more faithfully capture true object structure.
    }
    \label{fig:qual_results}
\end{figure*}

\paragraph{Quantitative results}
We present quantitative evaluations across the Mip‑NeRF360~\cite{mipnerf360}, LERF Mask~\cite{gaussian_grouping}, and LLFF~\cite{llff} datasets in~\cref{tab:quant_res}. 
Our approach achieves competitive or superior mIoU relative to state‑of‑the‑art Gaussian‑based methods. 
While mean accuracy (mAcc) is on par with existing baselines, the discrepancy -- higher mIoU with comparable mAcc -- highlights a critical issue inherent in baseline methods. 
Baseline Gaussian‑based methods often suffer from \ul{over‑segmentation}, which artificially inflates accuracy scores while substantially degrading the intersection‑over‑union (see \cref{fig:qual_results}). 
In contrast, our model maintains precise boundaries and avoids spurious fragments, leading to more stable and reliable segmentation. 
Furthermore, Semantic Foam retains the high reconstruction fidelity characteristic of the underlying Radiant Foam representation; detailed reconstruction metrics are provided in the supplementary material.

\paragraph{Qualitative results}
\Cref{fig:qual_results} illustrates the quality of our extracted assets and corresponding segmentation maps. 
While prior methods frequently produce dilated segmentations that incorporate background regions, our approach yields clean, well‑localized masks, enabling high‑fidelity object extraction. 
Additional qualitative comparisons, including video demonstrations of side‑by‑side asset extractions along novel camera trajectories, are available at \url{http://semanticfoam.github.io/} 

\begin{figure*}
    \centering
    \begin{overpic}[width=0.32\linewidth]{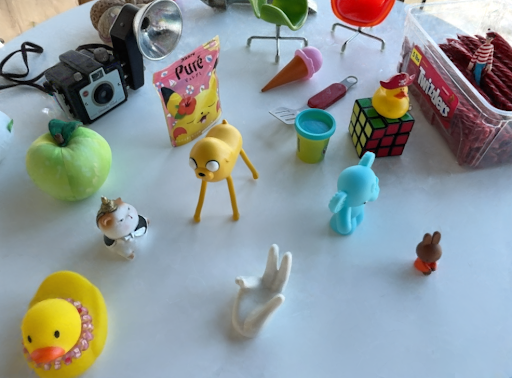}
    \end{overpic}
    \begin{overpic}[width=0.32\linewidth]{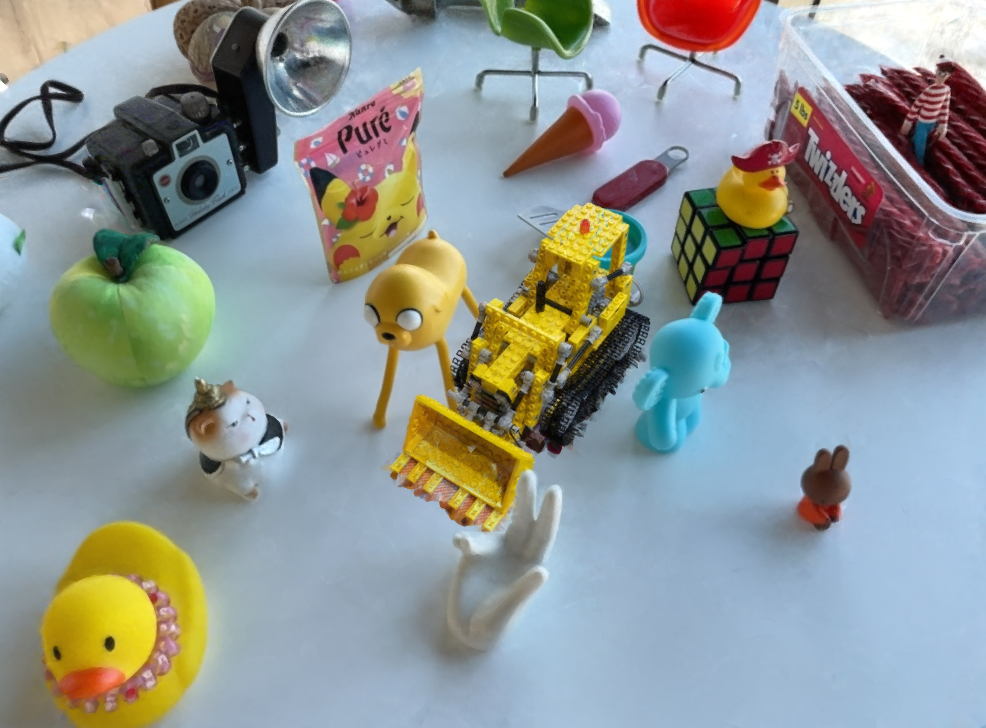}
    \end{overpic}
    \begin{overpic}[width=0.32\linewidth]{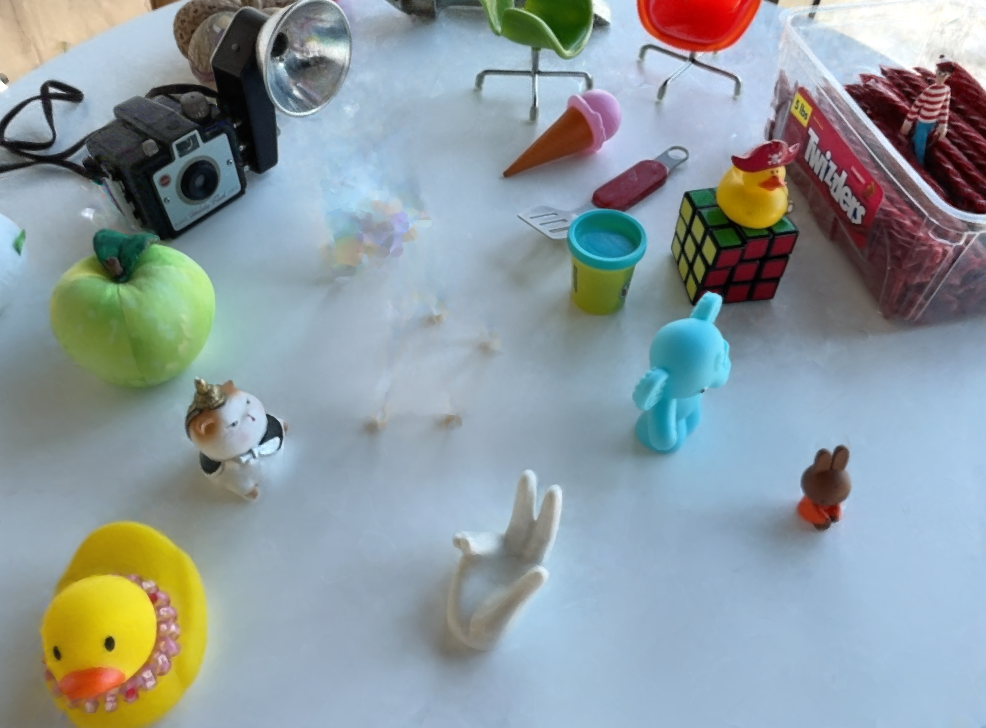}
    \end{overpic}
    \caption{{\bf Scene editing -- }
    We demonstrate our method's ability to edit scenes by insertion (middle), and deletion (right) of objects, with the (left) view showing the unedited reference image.
    Here, the original scene is the Figurines sequence from LERF-Masked~\cite{lerf}, while the inserted object is from the Kitchen scene in Mip-NeRF 360~\cite{mipnerf360}.
    This is achieved using the learned identity features, and allows for moving objects between different scenes.
    }
    \label{fig:editing}
\end{figure*}

\begin{figure}
    \centering
    \begin{overpic}[width=\linewidth]{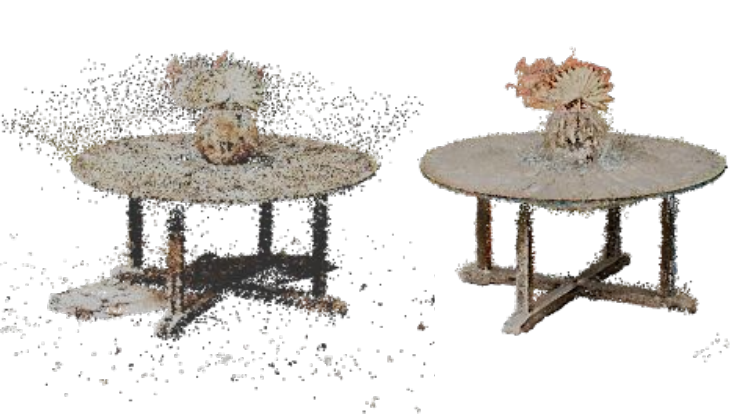}
    \put(74, 0){Ours}
    \put(7.5, 0){Gaussian Grouping~\cite{gaussian_grouping}}
    \end{overpic}
    \caption{
    {\bf Object extraction comparison -- }
    We demonstrate the extraction of the table and pot from the Garden scene~\cite{mipnerf360} through a corresponding point‑cloud visualization. 
    Unlike Gaussian Grouping -- which is restricted by convex-hull–based extraction -- our representation successfully handles a broad range of object geometries, including highly non-convex shapes, where Gaussian Grouping consistently fails.
    }
    \label{fig:gg_failure}
\end{figure}

\begin{figure}
    \centering
    \begin{overpic}[width=0.8\linewidth]{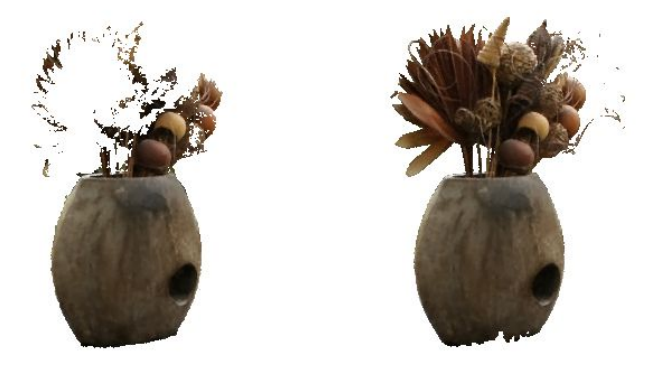}
    \put(13, -2){W/O $\losst{TV}$}
    \put(70, -2){W $\losst{TV}$}
    \end{overpic}
    \vspace{2pt}
    \caption{{\bf Ablation -- }
    We qualitatively assess the influence of the Total Variation loss on our method by examining the object mask of the pot from the Garden scene~\cite{mipnerf360}. 
    Without this regularization, the model exhibits a substantial decline in segmentation quality, producing object masks that fail to capture the complete object.
    }
    \label{fig:ablation}
\end{figure}

\subsection{Ablations}
\begin{table}[t]
\centering
\resizebox{\linewidth}{!}{
\setlength{\tabcolsep}{2pt}
\begin{tabular}{l|cc|cc|cc|cc}
    \toprule
    & \multicolumn{2}{c|}{Garden~\cite{mipnerf360}} 
    & \multicolumn{2}{c|}{Figurines~\cite{gaussian_grouping}} 
    & \multicolumn{2}{c|}{Horns~\cite{llff}}
    & \multicolumn{2}{c}{Average} \\
    & mIoU$\uparrow$ & mAcc$\uparrow$ 
    & mIoU$\uparrow$ & mAcc$\uparrow$ 
    & mIoU$\uparrow$ & mAcc$\uparrow$
    & mIoU$\uparrow$ & mAcc$\uparrow$ \\
    \midrule 
    W/O $\losst{TV}$ & 0.86 & 0.88 & \textbf{0.81}& \textbf{0.90}& \textbf{0.86} & \textbf{0.99} & 0.84 & 0.93 \\
    W $\losst{TV}$   & \textbf{0.94} & \textbf{0.96} & 0.79& 0.88& \textbf{0.86} & \textbf{0.99} & \textbf{0.86} & \textbf{0.94} \\
    \bottomrule
\end{tabular}
}
\caption{
    {\bf Ablation --}
    We evaluate the impact of Total Variation loss by excluding it and analyzing the semgnetation quality on the Garden scene from MipNeRF 360~\cite{mipnerf360}, Figurines scene from LERF-Masked~\cite{lerf}, as well as the Horns scene from LLFF~\cite{llff}.
}
\label{tab:ablation}
\end{table}
\paragraph{Total Variation Loss}
We conduct an ablation study evaluating our model with and without the Total Variation~(TV) loss across three representative scenes: Garden from Mip‑NeRF360~\cite{mipnerf360}, Figurines from LERF Masks~\cite{lerf}, and Horns from LLFF~\cite{llff}. 
We notice that incorporating the TV loss yields \textit{noticeably cleaner} and more structured identity encodings, which in turn enhances the fidelity of the resulting segmentation maps (see~\cref{fig:ablation}). 
This improvement is reflected in the superior performance in mIoU and mAcc reported for the full model in~\cref{tab:ablation}.

\subsection{Applications}
We evaluate the practical utility of Semantic Foam by demonstrating its capabilities for scene editing, specifically focusing on object extraction, removal, and insertion within complex 3D environments.

\paragraph{Object Extraction}
The 3D-consistent segmentation field learned by our model allows for robust object extraction \textit{directly} from the scene. 
Due to the inherent volumetric nature and implicit surface formulation of Radiant Foam, this process -- detailed in \cref{subsec:editing} -- is entirely seamless. 
Unlike Gaussian-based methods such as Gaussian Grouping~\cite{gaussian_grouping}, which necessitate \textit{post-hoc} heuristics like convex-hull construction to handle unstructured primitives, our method extracts assets natively. 
This leads to markedly cleaner results and higher fidelity to the original geometry, as evidenced by the comparisons in~\cref{fig:gg_failure}.

\paragraph{Object Removal}
Object removal leverages the same underlying principles of semantic and spatial decomposition. 
By identifying and pruning the relevant Voronoi cells as described in \cref{subsec:editing}, we eliminate target objects without introducing structural artifacts or requiring auxiliary cleanup steps. 
The effectiveness of this process in maintaining scene integrity is qualitatively demonstrated in~\cref{fig:editing}.

\paragraph{Object Insertion}
Finally, our representation supports the coherent insertion of novel assets. 
By localizing the insertion region and re-triangulating the Voronoi mesh to accommodate new primal sites~(\cref{subsec:editing}), we integrate external objects into the scene while preserving overall geometric consistency and photorealism. 
Visual results of these compositions are provided in~\cref{fig:editing}.

\section{Conclusion}
\vspace{-0.6em}
We propose a method for learning semantically decomposed Radiant Foam models which combine the inherent spatial decomposition of the foam model with a semantic decomposition that enables practical editing tasks like object extraction, deletion, and insertion.
Our method matches or outperforms semantically decomposed Gaussian Splatting baselines on segmentation tasks and avoids limitations and artifacts caused by the unstructured nature of Gaussian representations.
\vspace{-0.6em}

\paragraph{Limitations}
Our pipeline, similar to other 3D inverse rendering segmentation approaches, relies on supervision from models operating on 2D images.
This can result in inconsistent supervision signals across different images in the dataset, as the predictions are not natively 3D-aware.
This could be addressed in future work by leveraging 3D-aware foundation models to produce the supervision signal conditioned on depth, or by modifying the loss function to increase robustness to inconsistent supervision.

Another limitation is that our 3D segmentation process is evaluated \textit{per-primitive}, while the supervision during training is applied on features that have been accumulated through volume rendering.
This may be less than optimal, and future work could investigate ways to ensure alignment of the per-primitive and accumulated features.
\vspace{-0.6em}
\paragraph{Acknowledgements}
This work was supported in part by the Natural Sciences and Engineering Research Council of Canada (NSERC) Discovery Grant, NSERC Collaborative Research and Development Grant, Google DeepMind, Digital Research Alliance of Canada, the Advanced Research Computing at the University of British Columbia, Microsoft Azure, and the SFU Visual Computing Research Chair program. We would like to thank Francis Engelman for his feedback and early research discussions.

\clearpage

{
    \small
    \bibliographystyle{ieeenat_fullname}
    \bibliography{main}
}

\newpage

\subsection{Additional implementation details}
\paragraph{Training} 
The training pipeline utilizes the Adam optimizer~\cite{adam}. Similar to Radiant Foam \cite{govindarajan2025radfoam}, we directly optimize per-point position, density, and view-dependent color, the latter being represented via spherical harmonics (SH) of degree three, in conjunction with identity encodings.
Optimization of point coordinates commences with an initial learning rate of $2e^{-4}$ and is annealed using a cosine schedule to a minimum rate of $2e^{-6}$.
The initial learning rates for point density and spherical harmonics are set to $1e^{-1}$ and $5e^{-3}$, respectively. 
These rates are also decayed via a cosine annealing schedule to a final rate equal to $0.1$ times the initial rate.
Consistent with Radiant Foam, we initially optimize only the zero-order (ambient) component of the SH coefficients. 
Optimization for the high-order coefficients is subsequently introduced after a warmup period, spanning the first 25\% of the total training iterations.
For identity encodings, we start with a learning rate of $5e^{-3}$ and decay it to a final learning rate of $5e^{-4}$ following the same cosine annealing schedule.

Following Radiant Foam, once initialization and warm-up training are complete, we progressively increase the number of Voronoi sites, linearly expanding the point set until the target resolution is reached. 
The adjacency data structure is updated using the same schedule to ensure consistency throughout training. 
All experiments are run for 20k iterations, with the final 2k iterations refining only the radiance and density attributes while keeping point positions fixed.

For the total variation loss, we halt gradient propagation through the face area and clamp this value to a minimum of $1$, ensuring a stable and consistently non‑zero penalty across adjacent points. 
We likewise stop density gradients from flowing through the identity loss to avoid unintended density variations arising from noise in the segmentation masks. 
This strategy preserves the structural and geometric fidelity learned from image renderings while enabling optimization of identity encodings for scene semantics.

\subsection{Reconstruction metrics}
\begin{table}[t]
\centering
\resizebox{\linewidth}{!}{
\setlength{\tabcolsep}{2pt}
\begin{tabular}{l|ccc|ccc|ccc}
    \toprule
    & \multicolumn{3}{c|}{MipNeRF 360~\cite{mipnerf360}} & \multicolumn{3}{c|}{LERF-Masked~\cite{gaussian_grouping}} & \multicolumn{3}{c}{LLFF~\cite{llff}} \\
    & PSNR$\uparrow$ & SSIM$\uparrow$ & LPIPS$\downarrow$ & PSNR$\uparrow$ & SSIM$\uparrow$ & LPIPS$\downarrow$ & PSNR$\uparrow$ & SSIM$\uparrow$ & LPIPS$\downarrow$ \\
    \midrule 
    Radiant Foam  & 29.92 & 0.83 & 0.21 & 22.73 & 0.79 & 0.38 & 24.60 & 0.74 & 0.34\\
    Semantic Foam & 29.79 & 0.90 & 0.17 & 22.72 & 0.79 & 0.38 & 24.59 & 0.74 & 0.34\\
    \bottomrule
\end{tabular}
}
\caption{
    {\bf Reconstruction metrics --}
    We show that we maintain the reconstruction quality of Radiant Foam with our Semantic Foam model while simultaneously learning semantic segmentation. 
}
\label{tab:psnr_comp}
\end{table}

In ~\cref{tab:psnr_comp}, we compare our reconstruction metrics against the original Radiant Foam. 
Our results demonstrate that the proposed method preserves reconstruction fidelity to a nearly identical degree while simultaneously enabling the learning of semantic information.

\subsection{Per scene metrics}
\Cref{tab:mipnerf_all,tab:lerf_all,tab:llff_all} summarize the segmentation metrics collected for our evaluation of all considered techniques. 
These include results for Mip-NeRF360~\cite{mipnerf360}, LERF-Masked~\cite{gaussian_grouping} and LLFF~\cite{llff} scenes. 

\begin{table}[t]
\centering
\resizebox{\linewidth}{!}{
\setlength{\tabcolsep}{4pt}
\begin{tabular}{l|cccc}
    \toprule
     & LabelGS~\cite{zhang2025labelgs} & Gaussian Gouping~\cite{gaussian_grouping} & SAGA~\cite{cen2023saga} & SemanticFoam \\
     & mIoU$\uparrow$ / mAcc$\uparrow$ & mIoU$\uparrow$ / mAcc$\uparrow$ & mIoU$\uparrow$ / mAcc$\uparrow$ & mIoU$\uparrow$ / mAcc$\uparrow$ \\
    \midrule
    Garden  & 0.79 / 0.95 & 0.88 / 0.92& 0.78 / 0.94 & \textbf{0.94} / \textbf{0.96} \\
    Bonsai  & 0.70 / 0.92& 0.75 / 0.82& 0.96 / \textbf{0.97}& \textbf{0.90} / 0.94 \\
    Room    & 0.64 / 0.93 & 0.55 / 0.78& \textbf{0.80} / \textbf{0.99}& 0.63 / 0.95\\
    Counter & 0.55 / 0.93 & 0.74 / 0.92 & \textbf{0.81} / \textbf{0.96} & 0.75 / 0.91 \\
    Kitchen & 0.82 / 0.96& 0.83 / 0.92& \textbf{0.95} / \textbf{0.99} & 0.90 / 0.94\\
    \bottomrule
\end{tabular}
}
\caption{
    {\bf Per-scene metrics --}
    Per-scene metrics – mIoU, and mAcc scores for MipNerf360~\cite{mipnerf360} scenes.
}
\label{tab:mipnerf_all}
\end{table}

\begin{table}[t]
\centering
\resizebox{\linewidth}{!}{
\setlength{\tabcolsep}{4pt}
\begin{tabular}{l|cccc}
    \toprule
     & LabelGS~\cite{zhang2025labelgs} & Gaussian Gouping~\cite{gaussian_grouping} & SAGA~\cite{cen2023saga} & SemanticFoam \\
     & mIoU$\uparrow$ / mAcc$\uparrow$ & mIoU$\uparrow$ / mAcc$\uparrow$ & mIoU$\uparrow$ / mAcc$\uparrow$ & mIoU$\uparrow$ / mAcc$\uparrow$ \\
    \midrule
    Figurines & 0.60 / 0.94& 0.75 / 0.91& \textbf{0.86} / \textbf{0.96}& 0.79 / 0.88\\
    Ramen & 0.39 / 0.97& \textbf{0.81} / 0.94& 0.54 / \textbf{0.99} & 0.79 / 0.87\\
    Teatime & 0.64 / 0.94& \textbf{0.76} / 0.81& 0.73 / \textbf{0.97}& \textbf{0.76} / 0.88\\
    \bottomrule
\end{tabular}
}
\caption{
    {\bf Per-scene metrics --}
    Per-scene metrics – mIoU, and mAcc scores for LERF-Masked~\cite{gaussian_grouping} scenes.
}
\label{tab:lerf_all}
\end{table}

\begin{table}[t]
\centering
\resizebox{\linewidth}{!}{
\setlength{\tabcolsep}{2pt}
\begin{tabular}{l|cccc}
    \toprule
     & LabelGS~\cite{zhang2025labelgs} & Gaussian Grouping~\cite{gaussian_grouping} & SAGA~\cite{cen2023saga} & SemanticFoam \\
     & mIoU$\uparrow$ / mAcc$\uparrow$ & mIoU$\uparrow$ / mAcc$\uparrow$ & mIoU$\uparrow$ / mAcc$\uparrow$ & mIoU$\uparrow$ / mAcc$\uparrow$ \\
    \midrule
    Fern     & 0.85 / \textbf{1.00} & \textbf{0.93} / \textbf{1.00} & 0.90 / 0.98 & 0.90 / 0.98 \\
    Flower   & 0.70 / 0.70 & 0.73 / 0.79 & \textbf{0.95} / \textbf{1.00} & 0.73 / 0.79 \\
    Fortress & 0.70 / \textbf{1.00} & 0.98 / \textbf{1.00} & \textbf{0.99} / \textbf{1.00} & 0.98 / \textbf{1.00} \\
    Horns    & 0.92 / 0.92 & 0.95 / \textbf{0.99} & \textbf{0.96} / 0.98 & 0.95 / 0.98 \\
    Leaves   & 0.02 / 0.03 & 0.92 / 0.99 & 0.49 / 0.95 & \textbf{0.97} / \textbf{1.00} \\
    Orchids  & \textbf{0.82} / \textbf{0.95} & 0.87 / \textbf{0.96} & 0.36 / 0.38 & 0.78 / 0.91 \\
    Room     & 0.87 / \textbf{0.99} & \textbf{0.97} / \textbf{0.99} & \textbf{0.97} / 0.98 & 0.95 / 0.95 \\
    Trex     & 0.44 / \textbf{0.61} & 0.54 / \textbf{0.58} & 0.21 / 0.21 & \textbf{0.55} / 0.57 \\
    \bottomrule
\end{tabular}
}
\caption{
    {\bf Per-scene metrics --}
    Per-scene metrics – mIoU, and mAcc scores for LLFF~\cite{llff} scenes.
}
\label{tab:llff_all}
\end{table}

\subsection{Training time and memory consumption}
In~\cref{tab:time_metrics}, we report the average training time, inference speed, and memory usage of Semantic Foam compared to Radiant Foam, evaluated on the LERF-Masked~\cite{llff} dataset using an NVIDIA RTX A6000 GPU. 
Our model preserves the training and inference efficiency of Radiant Foam, while incurring an 18\% increase in model size due to the inclusion of identity features required for the segmentation task. 
\begin{table}[h]
\centering
\resizebox{\linewidth}{!}{
\setlength{\tabcolsep}{2pt}
\begin{tabular}{lcccccc}
    \toprule
     & \multicolumn{2}{c}{Training Time (mins $\downarrow$)} 
     & \multicolumn{2}{c}{Model Size (MB $\downarrow$)} 
     & \multicolumn{2}{c}{Inference Speed (FPS $\uparrow$)} \\
     \cmidrule(lr){2-3} \cmidrule(lr){4-5} \cmidrule(lr){6-7}
     & Radiant Foam & Semantic Foam & Radiant Foam & Semantic Foam & Radiant Foam & Semantic Foam \\
    \midrule
    Figurines & 83.00 & 84.00 & 661.00 & 783.00 & 84.52 & 84.17 \\
    Ramen & 80.00 & 80.00 & 655.00 & 778.00 & 59.75 & 59.79 \\
    Teatime & 79.00 & 79.00 & 663.00 & 785.00 & 91.04 & 90.39 \\
    \midrule
    Average & 80.67 & 81.00 & 659.67 & 782.00 & 78.44 & 78.11 \\
    \bottomrule
\end{tabular}
}
\caption{
    {\bf Model and training statistics}
}
 \label{tab:time_metrics}
\end{table}

\subsection{Additional Quantitative Comparisons} 
In this section, we compare our model with InstanceGaussian and GARField.
InstanceGaussian is designed primarily for open-language grounded evaluation, which differs fundamentally from the click-based object evaluation framework utilized by our method and existing baselines. 
We adapt the evaluation protocol from InstanceGaussian to facilitate a direct comparison on the LERF-Masked dataset, as shown in~\cref{tab:add_exps}. 
We also evaluate our method against GARField -- a recent state-of-the-art NeRF-based semantic segmentation method -- on the LERF-Masked dataset. 
As detailed in~\cref{tab:add_exps}, our method surpasses InstanceGaussian and GARField in average mIoU metrics.
\begin{table}[h]
\centering
\resizebox{\linewidth}{!}{
\setlength{\tabcolsep}{4pt}
\begin{tabular}{l|ccc|c}
    \toprule
     & Figurines & Ramen & Teatime & Average \\
    \midrule
    InstanceGaussian 
        & 0.63/\textbf{0.96}& 0.48/\textbf{0.89}& 0.73/\textbf{0.96}& 0.61/\textbf{0.96}\\
    GARField 
        & 0.65/0.70& 0.49/0.57& 0.63/0.66& 0.59/0.68\\
    Ours 
        & \textbf{0.79}/0.94& \textbf{0.79}/0.87& \textbf{0.76}/0.88& \textbf{0.79}/0.88\\
    \bottomrule
\end{tabular}
}
\caption{
    {\bf Additional quantitative comparisons (mIoU$\uparrow$/mAcc$\uparrow$)}
}
\label{tab:add_exps}
\end{table}

\subsection{Additional qualitative results}
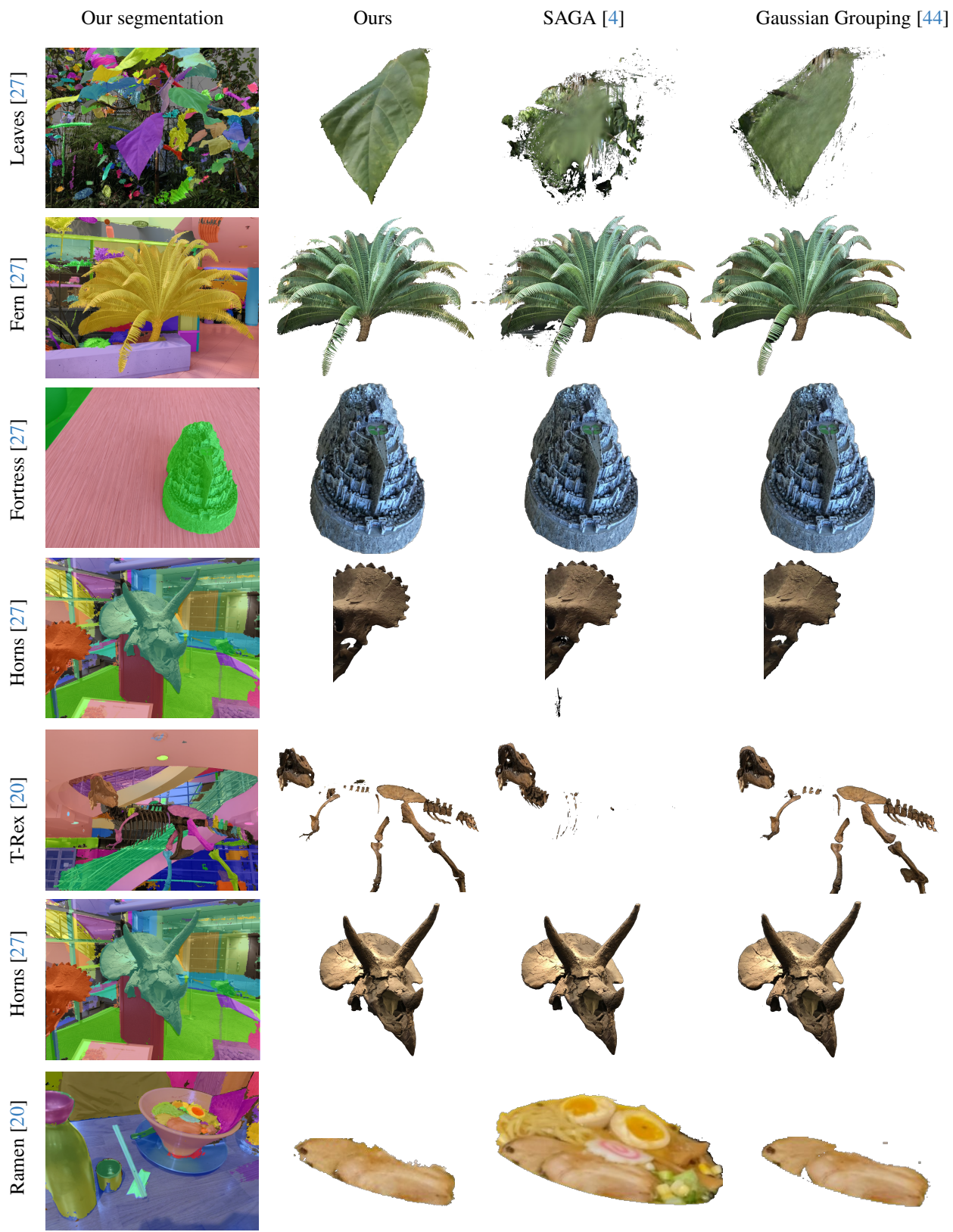
\begin{figure*}
    \centering
    \begin{overpic}[width=0.9\linewidth]{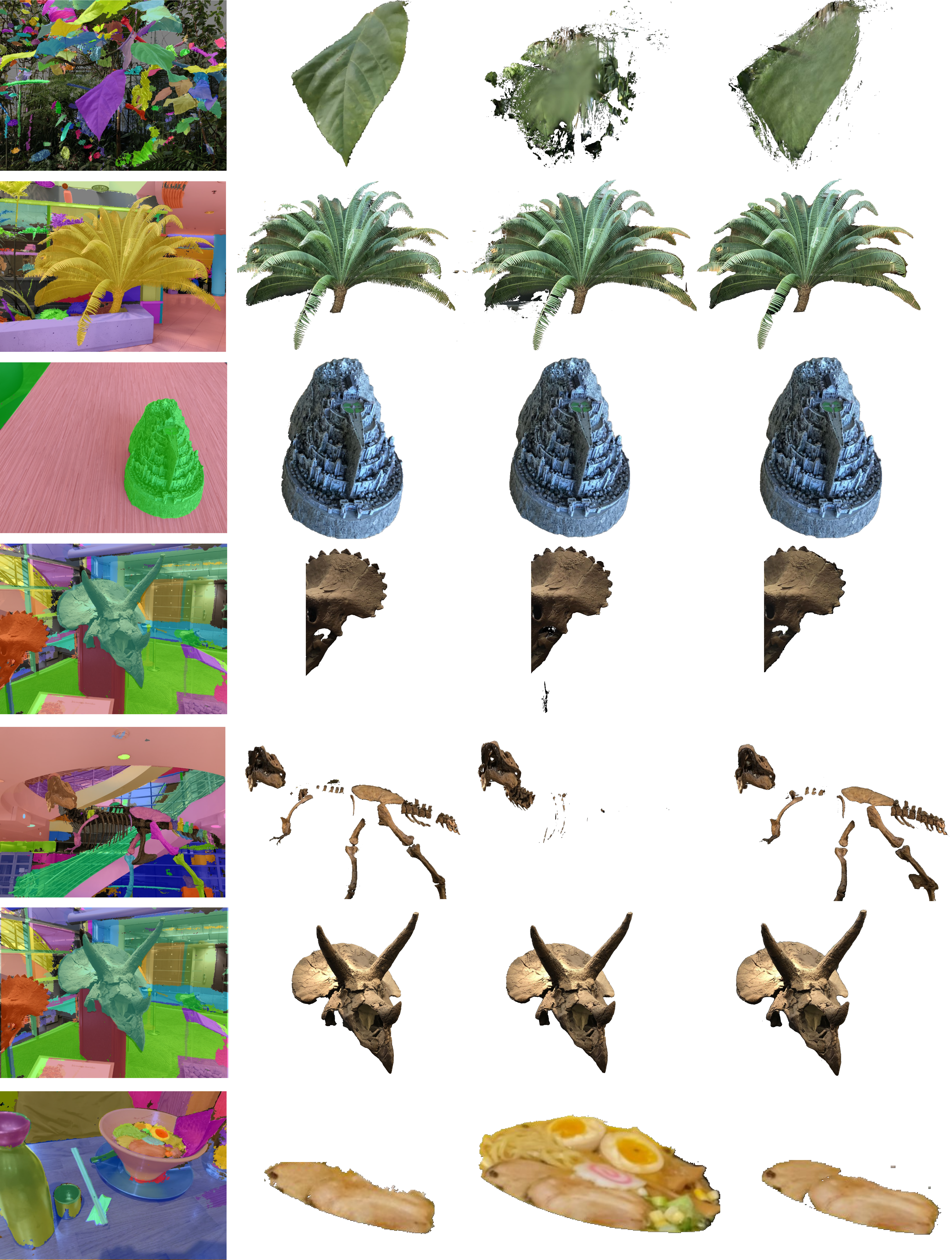}
    \put(-3, 90){\rotatebox{90}{Leaves~\cite{llff}}}
    \put(-3, 76){\rotatebox{90}{Fern~\cite{llff}}}
    \put(-3, 60){\rotatebox{90}{Fortress~\cite{llff}}}
    \put(-3, 46){\rotatebox{90}{Horns~\cite{llff}}}
    \put(-3,31){\rotatebox{90}{T-Rex~\cite{lerf}}}
    \put(-3, 18){\rotatebox{90}{Horns~\cite{llff}}}
    \put(-3, 3){\rotatebox{90}{Ramen~\cite{lerf}}}

    \put(60, 102){Gaussian Grouping~\cite{gaussian_grouping}}
    \put(42,102){SAGA~\cite{cen2023saga}}
    \put(26,102){Ours}
    \put(3, 102){Our segmentation}
    \end{overpic}
    \caption{
    {\bf Extra Qualitative Results.}
    We present additional qualitative comparisons of our object - extraction results against Gaussian Grouping~\cite{gaussian_grouping} and SAGA~\cite{cen2023saga}.
    As demonstrated by the leaves and teatime extractions, Gaussian-based baselines often exhibit inconsistent segmentation boundaries; conversely, our approach generates sharp, accurately bounded masks that more faithfully preserve the integrity of the object's geometric structure.
    }
    \label{fig:extra_qual_results}
\end{figure*}

In~\cref{fig:extra_qual_results}, we present additional qualitative results that further illustrate the high fidelity of our extracted assets and their corresponding segmentation maps. While prior methods frequently produce dilated segmentations that incorporate background regions, our approach yields clean, well-localized masks, enabling high-fidelity object extraction.

\subsection{Additional scene editing results}
\begin{figure*}
    \centering
    \begin{overpic}[width=\linewidth]{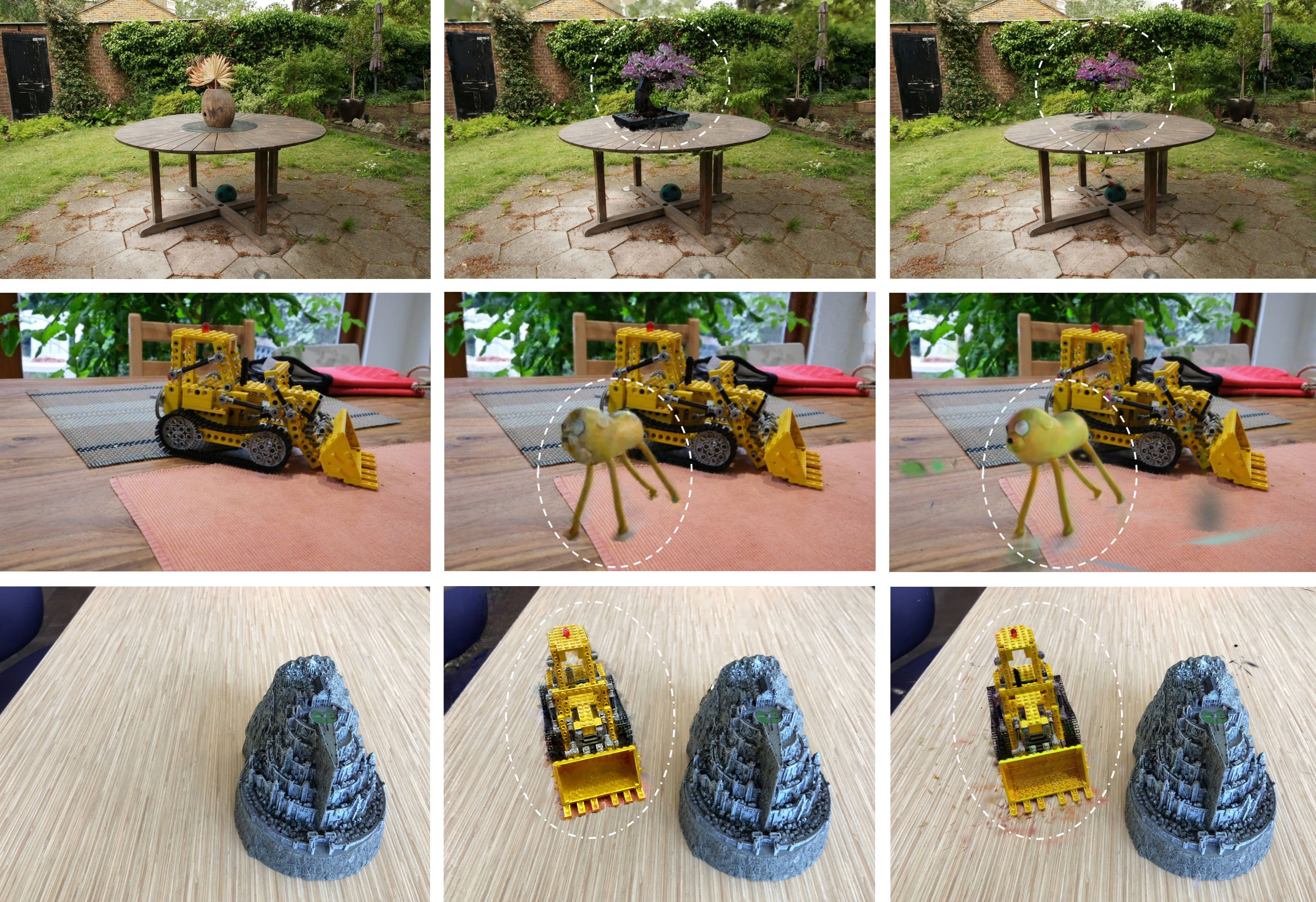}
    \put(11, -3){Unedited GT}
    \put(45, -3){Our insertion}
    \put(72, -3){Gaussian Grouping insertion}
    \end{overpic}
    \vspace{1pt}
    \caption{{\bf Scene editing (insertion) -- }
    Comparison of object insertion between our semantic foam representation (middle) and Gaussian Grouping (right), with the (left) view showing the unedited reference image. 
    Leveraging Radiant Foam’s implicit surface formulation, our method defines accurate non-convex 3D object masks without requiring convex-hull post-processing. 
    As shown, our approach cleanly inserts the toy and lego in the Kitchen and Fortress scenes respectively, whereas Gaussian Grouping inserts additional noise.
    }
    \label{fig:insertion_supp}
\end{figure*}

\begin{figure*}
    \centering
    \begin{overpic}[width=\linewidth]{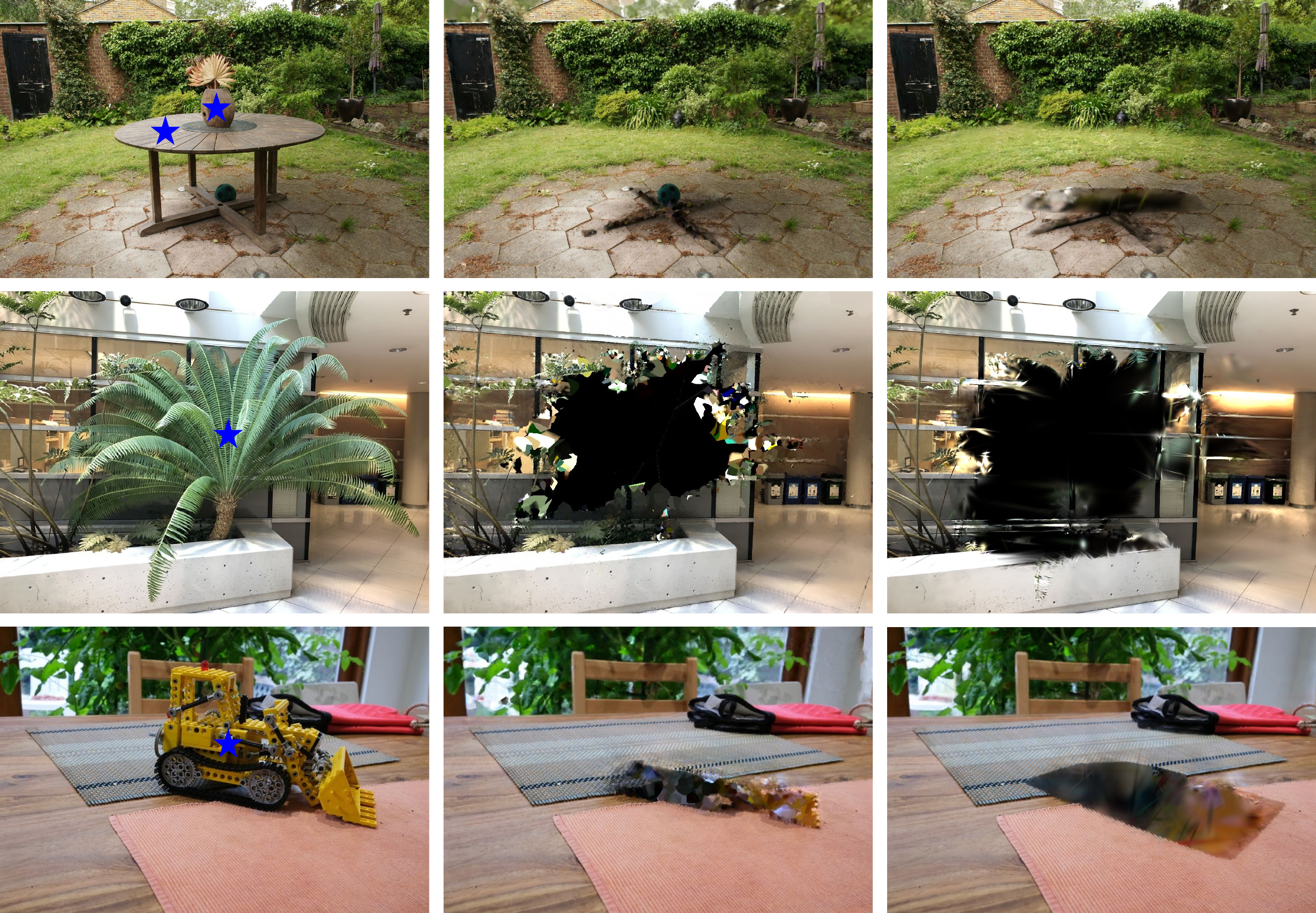}
    \put(11, -3){Unedited GT}
    \put(45, -3){Our deletion}
    \put(72, -3){Gaussian Grouping deletion}
    \end{overpic}
    \vspace{1pt}
    \caption{{\bf Scene editing (deletion) -- }
    Comparison of object deletion between our semantic foam representation (middle) and Gaussian Grouping (right), with the (left) view showing the unedited reference image (blue star denotes the object selected for deletion). 
    Leveraging Radiant Foam’s implicit surface formulation, our method defines accurate non-convex 3D object masks without requiring convex-hull post-processing. 
    As illustrated, our approach accurately isolates the table and pot in the Garden scene, while Gaussian Grouping over‑deletes and erroneously removes the nearby ball due to its convexity‑restricted mask formulation.
    }
    \label{fig:deletion_supp}
\end{figure*}

In~\cref{fig:insertion_supp,fig:deletion_supp}, we compare the object insertion and deletion capabilities of our semantic foam representation against Gaussian Grouping.
We restrict our comparison to Gaussian Grouping because SAGA and Label-GS do not provide open‑source code for object‑level editing.
Owing to Radiant Foam’s implicit surface formulation, our method enables seamless definition of 3D object masks without relying on additional post‑processing steps such as the convex‑hull construction required by Gaussian Grouping. 
As a consequence, Gaussian Grouping often over‑deletes or introduces noise into the scene, since its object boundaries are restricted to convex shapes. 
For instance, as shown in~\cref{fig:deletion_supp}, our model accurately extracts only the table and pot from the Garden scene, whereas Gaussian Grouping also removes the nearby ball due to its inability to represent non‑concave object masks.

\end{document}